\documentclass[10pt,twocolumn,letterpaper]{article}

\usepackage{cvpr}              

\usepackage{algorithm}
\usepackage{algcompatible}

\usepackage[accsupp]{axessibility}  

\usepackage[dvipsnames]{xcolor}

\definecolor{cvprblue}{rgb}{0.21,0.49,0.74}
\usepackage[pagebackref,breaklinks,colorlinks,citecolor=cvprblue]{hyperref}

\def\paperID{CLVision Workshop \#30} 
\def\confName{CVPR}
\def\confYear{2024}

\author{%
Haoran Zhu$^{1*}$\quad
Maryam Majzoubi $^{2*}$ \quad
Arihant Jain $^{1*}$ \quad
Anna Choromanska $^{1}$\vspace{0.2em}\\
$^{1}$New York University \quad $^{2}$Google \quad \quad \\
{\tt\small{\{hz1922, aj2622, ac5455\}@nyu.edu \quad maryam.majzoubi@gmail.com}} \\
}

\newcommand{\tame}{\ensuremath{\mathtt{TAME}}\xspace}
\newcommand{\addexpert}{\ensuremath{\mathtt{Add\_expert}}\xspace}
\newcommand{\getthresh}{\ensuremath{\mathtt{Get\_threshold}}\xspace}

\begin{document}

\title{TAME: Task Agnostic Continual Learning using Multiple Experts}

\maketitle
\begin{abstract}
The goal of lifelong learning is to continuously learn from non-stationary distributions, where the non-stationarity is typically imposed by a sequence of distinct tasks. Prior works have mostly considered idealistic settings, where the identity of tasks is known at least at training. In this paper we focus on a fundamentally harder, so-called task-agnostic, setting where the task identities are not known and the learning machine needs to infer them from the observations. Our algorithm, which we call TAME (\textbf{T}ask-\textbf{A}gnostic continual learning using \textbf{M}ultiple \textbf{E}xperts), automatically detects the shift in data distributions and switches between \textit{task expert networks} in an online manner. At training, the strategy for switching between tasks hinges on an extremely simple observation that for each new coming task there occurs a statistically-significant deviation in the value of the loss function that marks the onset of this new task. At inference, the switching between experts is governed by the \textit{selector} network that forwards the test sample to its relevant expert network. The selector network is trained on a small subset of data drawn uniformly at random. We control the growth of the task expert networks as well as selector network by employing pruning.
Our experimental results show the efficacy of our approach on benchmark continual learning data sets, outperforming the previous task-agnostic methods and even the techniques that admit task identities at both training and testing, while at the same time using a comparable model size.
\end{abstract}

\section{Introduction}

Learning agents deployed in real world applications are exposed to a continuous stream of incrementally available information usually from non-stationary data distributions. 
The agent is required to adaptively learn over time by accommodating new experience while preserving previous learned knowledge. This is referred to as lifelong or continual learning, which has been a long-established challenge in artificial intelligence, including deep learning~\cite{PARISI201954,Neuroscience-Inspired,THRUN199525}.

In the commonly considered scenario of lifelong learning, where the tasks come sequentially and each task is a sequence of events from the same distribution, one of the main challenges is to overcome catastrophic forgetting, where training the model on a new task interfere with the previously acquired knowledge and leads to the performance deterioration on the previously seen tasks. Deep neural networks generally perform well on classification tasks, but they heavily rely on having i.i.d. data samples drawn from stationary distribution during training time~\cite{GUO201627,article,maxtransfer}. In the case of sequential tasks, their performance significantly deteriorates when learning new coming tasks~\cite{PARISI201954,McClelland,McCloskey,kemker,Maltoni}.

A number of approaches have been suggested in the literature to deal with catastrophic forgetting.
Some works~\cite{hsu2019reevaluating,vandeven2019generative} provide systematic
categorization of the continual learning frameworks and identify three different scenarios: incremental task learning, incremental domain learning, and incremental class learning, where their differences stem from the availability of task labels at testing and number of output heads. In incremental class and domain learning the task identity is not known during the testing. All of these scenarios however are based on the assumption that the task labels are known at the training phase. This assumption is limiting in practical real-world applications, where the agent needs to learn in a more challenging task-agnostic setting~\cite{bgd,curl,itaml,cndpm}. In this learning setting the task identities are not available both at training and inference times. The literature started exploring this setting very recently and this setting is in the central focus of our paper.

In this work, we present an approach for handling task-agnostic continual learning inspired by the older approaches dedicated to 
learning non-stationary sequences based on experts advice~\cite{DBLP:journals/ml/HerbsterW98,monte1,monte2}, which explore and exploit the intermittent switches between distinct stationary processes. In these approaches the learner can make predictions on the basis of a fixed set of experts.
Since the learner does not
know the mechanisms by which the experts arrive at their predictions,
it ought to exploit the information obtained by observing the losses of
the experts. Based on the experts' losses it weights the experts to attenuate poor performers and emphasize the good ones, and forms the final prediction as the weighted sum of experts' predictions. Thus the learner needs to identify the best expert at each time and switch between the experts when the task switches occur. In the aforementioned works, the weights over the experts are the only carriers of the memory of previous experiences. Also, the discussed methods rely on the assumption that the number of experts/tasks are known in advance. Finally, these methods do not consider a separate train and test phase, but rather their optimization process is focused on minimizing the regret, which is the difference between the cumulative loss of the algorithm and the loss of the best method in the same class, chosen in hindsight (hindsight refers to full knowledge of the sequence to be predicted). Minimizing the regret however is not equivalent to counter-acting catastrophic forgetting since previous tasks that present little relevance to the currently learned ones are gradually being overwritten in memory. These methods thus are not directly applicable to the continual learning setting.

Motivated by having a set of experts representing a sequence of tasks, where each task is essentially a stationary segment of a longer non-stationary distribution, 

\begin{figure}[t]
  \centering
   \includegraphics[width=0.8\linewidth]{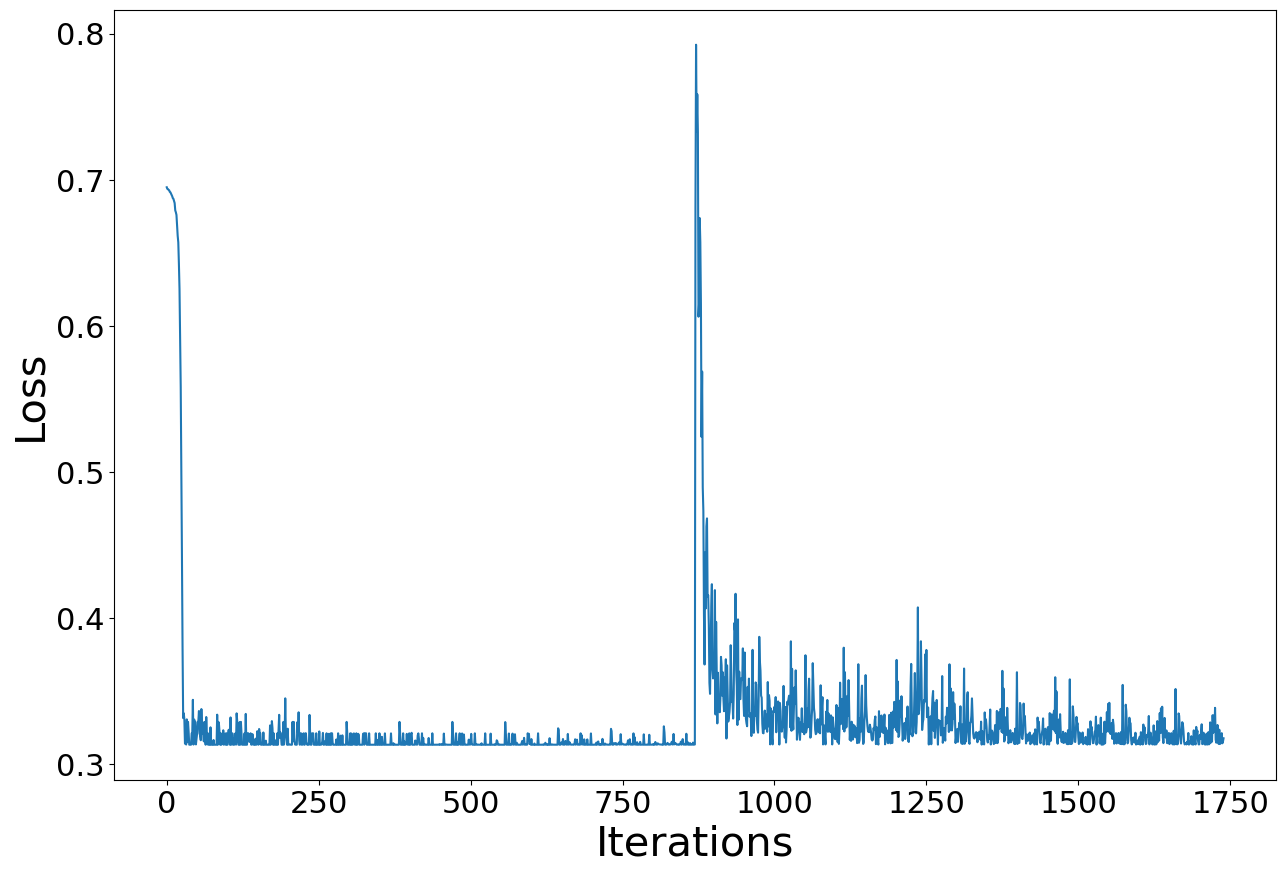}

   \caption{\label{fig:motiv}Deviation of the value of loss function of the expert when the task is switched}
\end{figure}

we propose a learning system that initially starts with one expert and gradually adds 
or switches between experts when the tasks change. During the online training phase our algorithm automatically identifies when the task switches and either selects or creates the best expert for a new task, depending whether this task was seen before or not. The detection of task switches relies on the statistically significant deviation of the loss function value of the current expert, which marks the onset of the new task (see Figure~\ref{fig:motiv}). Similarly, the determination whether the task was seen before or not relies on the behavior of the per-expert loss functions (if the deviation of all per-expert loss values are high, new expert is created to represent the current task). Such simple detection mechanism is inspired by the classical experts advise literature discussed in the previous paragraph, where switching between experts is governed by the values of the loss functions of the experts. Moreover, we introduce a \textit{selector} network which predicts the task identity of the samples at inference time. The selector network is trained on a small subset of training examples that were sampled uniformly at random from different tasks during the learning process. Despite the simplicity of our approach, it leads to a task-agnostic continual learning algorithm that compares favorably to existing methods and proves that a rich historical literature on online processing of non-stationary sequences can provide useful signal processing tools for addressing challenges in modern continual learning discipline. 

The rest of the paper is organized as follows: Section~\ref{sec:rel} discusses the most relevant work. Section~\ref{sec:alg} introduces our algorithm which we call TAME: \textbf{T}ask \textbf{A}gnostic continual learning using \textbf{M}ultiple \textbf{E}xperts. Section~\ref{sec:exp} reports empirical results on benchmark continual learning data sets, and finally Section~\ref{sec:con} concludes the paper.

\section{Related Work}
\label{sec:rel}

In recent years, there has been a plethora of techniques proposed for continual learning that mitigate the catastrophic forgetting problem in deep neural networks. The existing approaches can be divided into three categories: i) complementary learning systems and memory replay methods, ii) regularization-based methods, and iii) dynamic architecture methods. These techniques are not dedicated to the task-agnostic scenario since they assume the identity of the tasks are provided at least during the training phase. On the other hand, more challenging task-agnostic continual learning setting was addressed only recently in a handful of papers. We review them first since our paper considers the same setting. For completeness we also discuss the most relevant works from the broad continual learning literature and refer the reader to a survey paper~\cite{PARISI201954} that provides a more comprehensive review of these approaches.

\paragraph{Task-Agnostic Continual Learning} In the context of supervised learning setting, which is of central focus to this paper, one of the first methods addressing task-agnostic continual learning is the Bayesian Gradient Descent algorithm, popularly known as BGD~\cite{bgd}. This approach is based on an online version of variational Bayes and proposes a Bayesian learning update rule for the mean and variance of each parameter. As all Bayesian approaches, this method counter-acts catastrophic forgetting by using the posterior distribution of the parameters for the previous task as a prior for the new task. BGD obtains the most promising empirical results in the setting, where the method relies on the so-called ``label trick'' where the task identity is inferred from the class label. Label trick however breaks the task-agnostic assumption. Another approach called iTAML~\cite{itaml} proposes to use meta-learning to maintain a set of generalized parameters that represent all tasks. When presented with a continuum of data at inference, the model automatically identifies the task and quickly adapts to it with just a single update. However at training the inner loop of their algorithm, which generates task-specific models for each task that are then combined in the outer loop  to form a more generic model, requires the knowledge of task label. At inference, the task is predicted using generalized model parameters. Specifically, for each sample in the continuum, the outcome of the general model is obtained and a maximum response per task is recorded. An average of the maximum responses per task is used as the task score. A task with a maximum score is finally predicted. iTAML counter-acts catastrophic forgetting by keeping a memory buffer of samples from different tasks and using it to fine-tune generalized parameters representing all tasks to a currently seen one. This method is not task-agnostic, since it requires task labels at training, though the authors categorize their method as task-agnostic. CN-DPM~\cite{cndpm} is an expansion-based method that eliminates catastrophic forgetting by allocating new resources to learn new data. They formulate the task-agnostic continual learning problem as an online variational inference of Dirichlet process mixture models consisting of a set of neural experts. Each expert is in charge of a subset of the data. Each expert is associated with a discriminative model (classifier) and a generative model (density estimator). For a new sample, they first decide whether the sample should be assigned to an existing expert or a new expert should be created for it. This is done by computing the responsibility scores of the experts for the considered sample and is supported by a short-term memory (STM) collecting sufficient data. Specifically, when a data point is classified as new, they store it to the STM. Once the STM reaches its maximum capacity, they train a new expert with the data in the STM. Another technique for task-agnosic continual learning, known as HCL~\cite{hcl}, models the distribution of each task and each class with a normalizing flow model. For task identification, they use the state-of-the-art anomaly detection techniques based on measuring the typicality of the model’s statistics. For avoiding catastrophic forgetting they use a combination of generative replay and a functional regularization technique. 

In the context of unsupervised learning setting, VASE method~\cite{achille} addresses representation learning from piece-wise stationary visual data based on a variational autoencoder with shared embeddings. The emphasis of this work is put on learning shared representations across
domains. The method automatically detects shifts in the training data distribution and uses this information to
allocate spare latent capacity to novel data set-specific disentangled representations, while reusing
previously acquired representations of latent dimensions where applicable. Authors represent data sets using a set of data generative factors, where two data sets may use the same generative
factors but render them differently, or they may use a different subset of factors altogether. They next determine whether the average reconstruction error of the relevant generative factors for the current data matches the previous data sets by a threshold or not using Minimum Description
Length principle. Allocating spare representational capacity to
new knowledge protects previously learnt representations
from catastrophic forgetting. Another technique called CURL~\cite{curl} learns a task-specific representation on top of a larger set of shared parameters while dynamically
expanding model capacity to capture new tasks. The method represents tasks using a mixture of Gaussians and expands the model as needed, by maintaining a small set of poorly-modelled samples and then initialising and fitting a new mixture component to this set when it reaches a critical size. The method also relies on replay generative models to alleviate catastrophic forgetting.

\paragraph{Non Task-Agnostic Continual Learning} First family of non task-agnostic continual learning techniques consists of complementary learning systems and memory replay methods. They rely on replaying selected samples from the prior tasks. These samples are incorporated into the current learning process so that at each step the model is trained on a mixture of samples from a new task as well as a small subset of samples from the previously seen tasks. Some techniques focus on efficiently selecting and storing prior experiences through different selection strategies~\cite{selective,gradbased}. Other approaches, e.g. GEM~\cite{gem}, A-GEM~\cite{agem}, and MER~\cite{maxtransfer} focus on favoring positive backward transfer to previous tasks. Finally, there are deep generative replay approaches~\cite{deep-gen-replay,experience-replay} that substitute the replay memory buffer with a generative model to learn data distribution from previous tasks and generate samples accordingly when learning a new task. Another family of techniques, known as regularization-based methods, enforce a constraint on the parameter update of the neural network, usually by adding a regularization term to the objective function. This term penalizes the change in the model parameters when the new task is observed and assures they stay close to the parameters learned on the previous tasks. Among these techniques, we identify a few famous algorithms such as EWC~\cite{ewc}, SI~\cite{SI}, MAS~\cite{MAS}, and  RWALK~\cite{RWALK} that introduce different notions of the importance of synapses or parameters and penalizes changes to high importance parameters, as well as the LwF~\cite{lwf} method that can be seen as a combination of knowledge distillation and fine-tuning. Finally, the last family of techniques are the dynamic architecture methods that expand the architecture of the network by allocating additional resources, i.e., neurons or layers, to new tasks which is usually accompanied by additional parameter pruning and masking. This family consists of such techniques as expert-gate method~\cite{expert-gate}, progressive networks~\cite{progressive}, dynamically expandable network~\cite{den}, learn-to-grow method~\cite{learntogrow}, Packnet~\cite{packnet}, Piggyback~\cite{piggyback}, and hybrid schemes~\cite{compacting}. The last three techniques rely on network quantization and pruning to better control the complexity and size of the model. 

\section{TAME Algorithm}
\label{sec:alg}

In this section we describe the proposed algorithm \tame. Let $\mathcal{T}$ denote the set of all tasks. Each example is drawn i.i.d. from an unknown distribution $P_t$ of the corresponding task, i.e. $(x_i^t,y_i^t) \sim P_t$. The tasks come in a sequential manner. We consider a scenario where the task identity as well as the number of tasks is not known. The goal is to learn these tasks sequentially without catastrophic forgetting by automatically identifying the task identities both at the training as well as the testing time.

\tame is based on using multiple \textit{task expert networks}, where each expert network is associated with one task. At training, the algorithm automatically detects the shift in the data distribution in an online manner and switches between the existing experts or adds more experts if necessary. The strategy for the task switch detection relies on the statistically-significant deviation in the values of the loss function. At testing, we have an additional \textit{selector} network that automatically forwards each sample to its relevant expert. This selector network is trained on a small subset of samples that are drawn uniformly at random from the sequence of samples from all tasks.

\begin{figure}[t]
  \centering
   \includegraphics[width=0.8\linewidth]{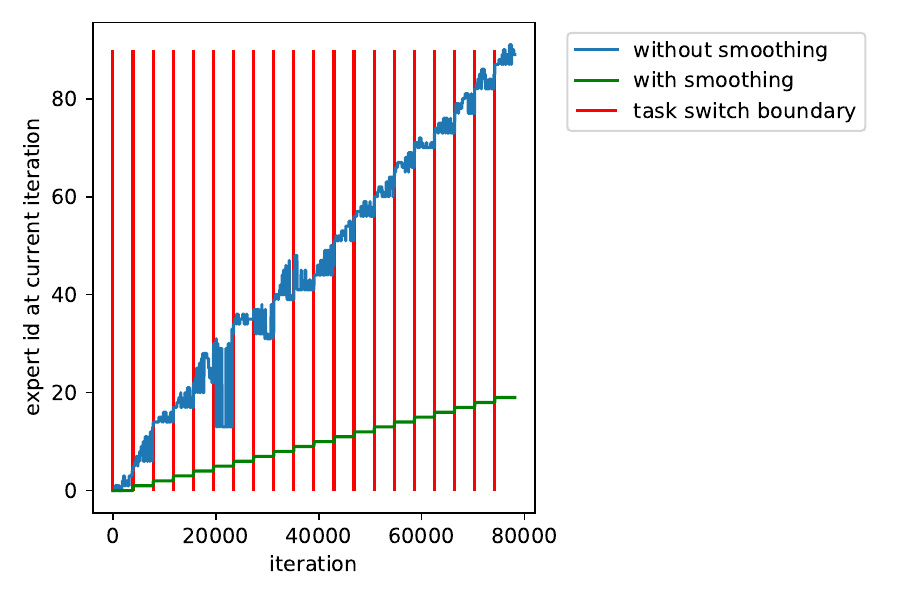}

   \caption{\label{fig:smoothing_effect} The effect of smoothing}
\end{figure}

The pseudo-code of our algorithm is captured in Algorithm~\ref{alg:tame}. We initially start with one expert network and gradually add more networks as needed. At each step of time we only have one active expert that is being trained on incoming data. We observe the value of the loss function of the current active expert. In order to smooth-out short term variations and highlight long-term patterns, a \textit{smoothed} version of the loss is calculated, through an exponentially weighted moving average (EWMA)~\cite{wiki:ma}. EWMA is a first-order infinite impulse response filter that applies weighting factors which decrease exponentially (never reaching zero). This is used to filter out higher frequency components that has no specific connection to shift in data distribution.
The coefficient $\alpha$ is a constant smoothing factor between $[0,1]$ that governs the amount of smoothing. Higher $\alpha$ diminishes previous observations faster. Figure \ref{fig:smoothing_effect} justifies the need for loss smoothing by comparing the performance of the proposed algorithm with and without smoothing. Smoothed loss helps avoiding false positives in detecting task switches, and thus prevents creating unnecessary experts, while at the same time it enables to maintain high detection accuracy.

Furthermore, we calculate a loss threshold value for each expert network. We assume a normal distribution for the value of the loss function. We set the significance threshold at three standard deviations above the mean. As shown in procedure \getthresh the mean and standard deviation are calculated over a moving window of size $W_{th}$ of the previously observed data. 

In lines $30-43$ we compare the smoothed loss with the threshold value of the current active expert and if it is above the threshold we search over all other existing experts and choose the one that meets the threshold requirement. If no such expert network is found, it means that no expert well-represents the currently seen data and thus a new expert is added to the model and gets activated (see procedure \addexpert). Next, the selected network is trained on the input data. 

We need to also train the selector network to switch between experts at inference. For this purpose, we have a buffer in the form of a priority queue with a fixed capacity $C_s$ that is much smaller than the total number of samples. In lines $46-50$, we randomly sub-sample data to keep it in the buffer in an online fashion. The label for each sample is the current expert id which corresponds to the task identity that we inferred from the data. The selector network is trained on the samples from this buffer and later used at inference time to automatically distinguish the task label and sends the test data to the corresponding expert network.

\begin{algorithm}[ht!]    
{\scriptsize
\caption{\tame: Task Agnostic continual learning using Multiple Experts}          
\label{alg:tame}                          
\begin{algorithmic} [1]                   
\REQUIRE{Data: $\{(x,y)\}$, Threshold window size: $W_{th}$, Smoothing factor: $\alpha$, Buffer capacity for training selector net: $C_s$, Buffer capacity for retraining after pruning: $C_p$}\\
\vspace{-0.05in}
\hrulefill\\
{\bf procedure: \addexpert()}\\
Input: $experts$, $N_e$\\
\quad Initialize a new $expert$ network.\\
\quad Initialize smoothed loss of the expert $expert.L_s \xleftarrow{} None$ \\
\quad Initialize $expert.deque$ with maximum capacity equal to $W_{th}$\\
\quad $expert.id = N_e$\\
\quad $experts$.add($expert$)\\
\quad $N_e \mathrel{+}= 1$\\
Return $expert$ \\

\hrulefill\\
{\bf procedure: \getthresh()}\\
Input: $expert.deque$\\
\quad $\mu$ = MEAN($expert.deque$)\\
\quad $\sigma$ = STD($expert.deque$)\\
Return $(\mu + 3*\sigma)$ \\

\hrulefill\\
\STATE {\bf Initialize}: buffer $buffer_{selector}$ with $C_s$ capacity; Buffers for pruning $buffers_{prune}  \xleftarrow{} []$ with $C_p$ capacity for all incoming tasks; number of expert: $N_e \xleftarrow{} 0$; $experts \xleftarrow{} []$; current task id $T_{id}\xleftarrow{} 0$;
\STATE $expert_c= \addexpert(experts,N_e)$
\STATE $T_{id} = 1$
\STATE $buffers_{prune}[T_{id}]  \xleftarrow{}$ a priority queue with $C_p$ capacity (initialize buffer for the first task)

\WHILE{Incoming Data}
\STATE $L_c = $ loss of the current expert $expert_c$ on input $\{(x,y)\}$
\label{line:smooth1}
\IF{$expert_c.L_s == None$}
\STATE $expert_c.L_s = L_c$
\ELSE
\STATE $expert_c.L_s = \alpha * L_c + (1 - \alpha) * expert_c.L_s$
\ENDIF
\label{line:smooth2}
\vspace{0.05in}
\IF{$expert_c.L_s > \getthresh(expert_c.deque)$}
\label{line:th1}
\STATE $expert_p = None$
\STATE $T_{id}\xleftarrow{} 0$
\FOR{$e$ in $experts$}
\STATE $T_{id}=T_{id}+1$
\IF{$\alpha * L_e + (1 - \alpha) * e.L_s < \getthresh(e.deque)$}
\STATE $expert_p = e$; {\bf break}
\ENDIF
\ENDFOR
\IF{$expert_p == None$}
\STATE $expert_c= \addexpert(experts,N_e)$
\STATE $buffers_{prune}[T_{id}+1] \xleftarrow{}$ a priority with $C_p$ capacity (initialize buffer for the new task)
\ENDIF
\ENDIF
\label{line:th2}
\vspace{0.05in}
\STATE Train $expert_c$ on batch of data $\{(x,y)\}$ and update its $deque$.\\
\vspace{0.05in}


\label{line:samp1}
\FOR{$(x_i,-)$ in $\{(x,y)\}$}
\STATE $priority = \mathcal{N}(0,1)$
\STATE $buffer_{selector}$.add(key:$priority$, value:$(x,expert_c.id)$)
\STATE $buffers_{prune}[T_{id}]$.add(key:$priority$, value:$(x,y)$)
\ENDFOR
\ENDWHILE
\vspace{0.05in}
\label{line:samp2}
\STATE Train and prune $selector$ network on samples in $buffer_{selector}$

\FOR{ i in \{1, 2, \dots, $N_e$\}}
\STATE Prune and retrain $experts[i]$ using buffer $buffers_{prune}[i]$ stored for $experts[i]$
\ENDFOR
\end{algorithmic}
}
\end{algorithm}

To reduce the size of the experts and selector network, we perform network pruning. Typically, after pruning, the model needs to be retrained to prevent drastic drop in performance. To enable retraining of experts we introduce set of buffers $buffers_{prune}$ to store samples for each expert (task). Each buffer is implemented as a priority queue with a fixed capacity $C_p$. When new expert is created, a buffer for that expert is added to the set. In lines $46-50$, we randomly sub-sample data and fill in the buffer in an online fashion. Thus, for each task we only keep a fixed amount of randomly selected samples. After training for all tasks is done, in line $52$, we prune and retrain the selector network using buffer $buffer_{selector}$. In lines $53-55$, we prune and retrain each expert using the corresponding buffer from $buffers_{prune}$.

\section{Experiments}
\label{sec:exp}
In this section we evaluate \tame on benchmark continual learning data sets and compare with other state-of-the-art methods, namely previously proposed task-agnostic methods: BGD~\cite{bgd}, iTAML \cite{itaml}, HCL \cite{hcl}, and CN-DPM \cite{cndpm}, as well as techniques that are not task-agnostic but are dedicated to the continual learning setting, such as DEN~\cite{den}, EWC~\cite{ewc}, SI~\cite{SI}, A-GEM~\cite{agem}, and RWALK~\cite{RWALK}. For evaluating the performance of the competitor algorithms we use open-source implementations, when available{\footnote{\url{https://github.com/facebookresearch/agem}(A-GEM, SI, RWALK, and EWC)}\footnote{\url{https://github.com/jaehong31/DEN} (DEN)} \footnote{\url{https://github.com/igolan/bgd/} (BGD)}\footnote{\url{https://github.com/brjathu/iTAML} (iTAML)} \footnote{\url{https://github.com/soochan-lee/CN-DPM} (CN-DPM)}}. 

\subsection{Data sets}
\sloppy We use standard continual learning data sets: (1) {\bf Permuted MNIST}, where a set of tasks is created by using a different random permutation of MNIST pixels. We generated a set of $20$ data sets accordingly that correspond to $20$ tasks. 
(2) {\bf Split MNIST}, where a set of tasks is constructed by taking pairs of digits from the original MNIST data set, i.e. $\mathcal{T} = \{\{0,1\},\{2,3\},\{4,5\},\{6,7\},\{8,9\}\}$. 
(3) {\bf Split CIFAR-100 (20)}, where the original CIFAR-100 is divided into $20$ disjoint subsets, each containing $5$ class labels, i.e. $\mathcal{T} = \{\{0-4\},\{5-9\},\dots,\{95-99\}\}$. 
For comparison with HCL method, we use additional data sets: (4) {\bf Split CIFAR-100 (10)}, where the original CIFAR-100 is divided into $10$ disjoint subsets, each containing $10$ class labels, i.e. $\mathcal{T} = \{\{0-9\},\{10-9\},\dots,\{90-99\}\}$. (5) {\bf Split CIFAR-10 (5)}, where the original CIFAR-10 is divided into $5$ disjoint subsets each, containing $2$ class labels, i.e. $\mathcal{T} = \{\{0-1\},\{2-3\},\dots,\{8-9\}\}$. (6) {\bf SVHN-MNIST}, that combines SVHN\cite{svhn} and MNIST data sets in a way that SVNH is the first task and MNIST is the second one. (7) {\bf MNIST-SVHN} where MNIST is the first task and SVNH is the second one.

We use the following size of the images in our data sets, i.e. $1\times 28\times 28$ for Split MNIST and Permuted MNIST, $3\times 32\times 32$ for Split CIFAR-100 (10), Split CIFAR-100 (20), and Split CIFAR-10 (5). For MNIST-SVHN and SVHN-MNIST, we upscale size of the images in MNIST to $3\times32\times32$ and  use the original image size of $3\times32\times32$ for SVHN data set. We normalize Split MNIST and Permuted MNIST data sets by mean $0.1307$ and standard deviation $0.3081$,  Split CIFAR-100 (10) and Split CIFAR-100 (20) by mean ($0.5071, 0.4867, 0.4408$) and standard deviation ($0.2675, 0.2565, 0.2761$), and Split CIFAR-10 (5) by mean ($0.5, 0.5, 0.5$) and standard deviation ($0.5, 0.5, 0.5$). For SVHN-MNIST and MNIST-SVHN data sets, we use mean mean $0.1307$ and standard deviation $0.3081$ for MNIST, and mean ($0.5, 0.5, 0.5$) and standard deviation ($0.5, 0.5, 0.5$) for SVHN. For Split CIFAR-100 (10) and Split CIFAR-100 (20), we use additional data augmentation techniques such as random crop, flip, and rotations. 

\subsection{Networks Architecture}
For all methods, except HCL, in case of Permuted-MNIST and Split MNIST data sets we use a network with $2$ convolutional layers, which are followed by the usual ReLU activation function, the max pooling operation, and fully connected layers for the task expert networks and we use a $2$-layer MLP for the selector network. For Split CIFAR-100 (20), we use a slightly modified version of VGG11~\cite{VGG} architecture for the task expert networks and a pre-trained ResNet18~\cite{resnet} for the selector network. The modification of VGG accommodates having $5$ outputs. To compare with HCL, we use the above mentioned $2$-layer convolutional network for SVHN-MNIST, MNIST-SVHN, and Split MNIST data sets. For Split CIFAR-10 (5) and Split CIFAR-100 (20) we use EfficientNet~\cite{efficientnet} model pretrained on ImageNet for the expert network and a pre-trained ResNet-18 for the selector network. Note that the choice of architectures we make is done on purpose to stay aligned with the architectures used by the competitor methods. In order to prevent the sudden jump of the loss function value during an initial stage of the training we add a sigmoid layer on the output of each model.

In competitor algorithms, for Permuted and Split MNIST data sets, we experimented with both the aforementioned $2$-layer convolutional network as well as a $2$-layer MLP and chose the best results. For Split CIFAR-100 (20), we used the VGG architecture except for BGD and CN-DPM. In the case of BGD, the VGG architecture led to the loss function divergence and unstable results, thus we used the architecture suggested by the authors in their paper for this data set. For CN-DPM the performance was worse when we used the VGG architecture so we report the results given in the CN-DPM's original paper. Furthermore, DEN implementation was not available for convolutional networks, so we were not able to test it for Split CIFAR-100 (20). For HCL, we report the results given in their paper on all listed data sets since their code is not publicly available.

\subsection{Training Details}
We train the models using SGD optimizer with learning rate equal to $0.1$, Nesterov momentum $0.9$, and weight decay $5e-4$ and the batch size of to $128$ for all data sets. We trained for 10 epochs on each task from Permuted MNIST and Split MNIST data sets and for 200 epochs for each task from Split CIFAR-100 (20). For Split CIFAR-100 (20) we drop the learning rate by the factor of $5$ at the 60th, 120th, and 160th epochs. We also apply a warm-up training in the first epoch to prevent the network divergence early in the training. We train 90 epochs for SVHN-MNIST and MNIST-SVHN, 15 epochs for Split CIFAR-10 (5) and Split CIFAR-100 (20). Finally, we use L1 unstructured pruning to reduce the model size for each expert and the expect selector network. We retrain each expert after pruning with the sub-sampled data stored during training. We use SGD optimizer with learning rate equal to $0.1$ and weight decay $1e-4$.

\subsection{Hyperparameters}
We next describe the values of the hyperparameters that are specific to our algorithm. The hyperparameters settings used in the experimets are summarized in Table~\ref{tab:hyper-tame}. In all experiments we use the same window size $W_{th}$ equal to $100$ and loss smoothing factor $\alpha$ of $0.2$.

For competitor algorithms we either used hyperparameter settings suggested by the authors or performed a parameter search. The scope of the hyperparameter search for the regularization-based methods and the training settings for A-GEM are shown in the supplementary materials.

\begin{table*}[h!]
    \begin{center}
    \caption{Hyperparameter settings used for \tame}
    \centering
    \scalebox{0.9}{
    \begin{tabular}{|c||c|c|c|}
    \hline 
    \text {\tame } & \text { Permuted MNIST } & \shortstack{\text { Split MNIST } \\ \text {  SVHN-MNIST} \\ \text {MNIST-SVHN}}&   \shortstack{\text {Split CIFAR-100 (20)}\\\text {Split CIFAR-100 (10)}\\\text {Split CIFAR-10 (5)}}\\
    \hline \text {Threshold window $W_{th}$ } & 100 & 100 & 100\\
    \hline \text { Smoothing factor $\alpha$ } & 0.2 & 0.2 & 0.2\\
    \hline
    \text { Buffer capacity $C_s$ } & 5000 & 2500 & 7500\\
    \hline
    \text { Buffer capacity $C_p$ } & 6000 & 1000 & 200\\
    \hline
    \text { Expert pruning rate (\%) } & 98 & 98 & 98\\
    \hline
    \text { Expert selector pruning rate (\%) } & 50 & 50 & 50\\
    \hline
    \hline
    \end{tabular}
    }
    \label{tab:hyper-tame}
    \end{center}
    \vspace{-0.5cm}
\end{table*}

\subsection{Results}
The metric we use for our evaluation is the average accuracy measured on the test set. The average accuracy is defined as $ACC = 1/T \sum_{i=1}^T R_{T,i}$, where $R_{T,i}$ is the classification accuracy of the model on task $i$, and $T$ is the number of tasks.

In Table~\ref{tab:acc} we compare the average accuracy obtained by \tame and other algorithms. For BGD method, we do not use any ``label trick'' approaches and thus run it in a purely task-agnostic setting, same as \tame, HCL, and CN-DPM. For iTAML, the task identity is known during training, thus it is not a  task-agnostic method under our standards, however since during inference they do not rely on the task identity we kept this method as our competitor. All the other algorithms have access to task descriptors both at training and testing. \tame achieves the highest accuracy and the smallest model size among all considered algorithms and data sets. Note that \tame also outperforms continual learning methods that have access to task labels at training and/or testing.  In Table~\ref{tab:acc_with_hcl} we compare the performance of \tame with HCL. Our method outperforms this approach as well.

\begin{table*}[ht!]
    \centering
    \caption{Average Accuracy (\%) obtained by \tame and other algorithms for Permuted MNIST, Split MNIST, and Split CIFAR-100}
    \scalebox{1.0}{
    \begin{tabular}{|c||l r|l r|l r|}
    \hline
    {Data sets (\#tasks)} &  \multicolumn{2}{c|}{Permuted MNIST (20)} &  \multicolumn{2}{c|}{Split MNIST (5)} &  \multicolumn{2}{c|}{Split
    CIFAR-100 (20)} \\
    & Acc. (\%) & Param. & Acc. (\%) & Param. &Acc. (\%) & Param.\\
    \hline
    \multicolumn{7}{|c|}{\bf\textit{task identity is known during training and testing}}\\
    \hline
    EWC   & 54.81 & 61.7K& 98.18 & 61.7K& 32.78 & 9.23M\\
    \hline
    SI    & 81.31 & 61.7K & 94.85 & 61.7K& 30.28 & 9.23M\\
    \hline
    A-GEM & 79.61 &  61.7K&97.72 &61.7K & 43.57 & 9.23M\\
    \hline
    RWALK & 46.23 & 61.7K & 96.84& 61.7K& 31.13& 9.23M\\
    \hline
    DEN & 83.61& 120.2K& 95.51 & 120.2K& NA&NA\\
    \hline
    \multicolumn{7}{|c|}{\bf\textit{task identity is known during training, but not during testing}}\\
    \hline
    iTAML & NA & NA & 97.95 & 61.7K & 54.55& 9.23M\\
    \hline
    \multicolumn{7}{|c|}{\bf\textit{task-agnostic (task id is not known during both training and testing)}}\\
    \hline
    BGD (without label trick) & 79.15&  61.7K&19.00  & 61.7K&3.77 &9.23M\\
    \hline
    CN-DPM & 14.99 & 616.1K & 94.19& 746.8K& 20.45 &19.20M\\
    \hline
    HCL & NA & NA & 90.89 &  NA & NA & NA\\
    \hline
    \tame & \textbf{87.32} & \textbf{55.53K}& \textbf{98.63}&\textbf{37.02K} &\textbf{62.39}&\textbf{9.02M}\\
    \hline
    \end{tabular}}
    \label{tab:acc}
\end{table*}

\begin{table*}[ht!]
    \centering
    \caption{Average Accuracy (\%) obtained by \tame and HCL}
    \scalebox{0.7}{
    \begin{tabular}{|c||c|c|c|c|c|}
    \hline
    {Data sets (\#tasks)} &  
    {SVHN-MNIST} & 
    {MNIST-SVHN} & 
    {Split MNIST (5)} & 
    {Split CIFAR-10 (5)} &
    {Split CIFAR-100 (10)} \\
     & Acc. (\%) & Acc. (\%) &Acc. (\%) &Acc. (\%) &Acc. (\%)\\
    \hline
    HCL-FR & 96.38 & 95.62 & 90.89 & 89.44 & 59.66\\
    \hline
    HCL-GR & 93.84 & 96.04 & 84.65 & 80.29 & 51.64 \\
    \hline
    \tame  & \bf 97.45 & \bf 97.63 & \bf 98.63 & \bf 91.32 & \bf 61.06 \\
    \hline
    \end{tabular}}
    \label{tab:acc_with_hcl}
\end{table*}

\begin{table*}[ht!]
    \centering
    \caption{Size $C_p$ of buffer used for retraining experts after pruning versus average accuracy for Split CIFAR-100 (20), Split MNIST, and Permuted MNIST data sets}
    \begin{tabular}{|c|c|c|c|c|c|c|c|c|c|}
        \hline
        Data sets  & 50 & 100 & 200 & 500 & 1000 & 2000 & 3000 & 6000\\
        \hline
        Split CIFAR-100 (20) &  56.12 & 57.86 & 62.39 & 63.47 & 64.41 & / & / & /\\
        \hline
        Split MNIST & 97.80 & 98.20 & 98.22 & 98.38  & 98.63 & 98.38 & / & /\\
        \hline
        Permuted MNIST & / &62.55 & 69.95 & 76.83 & 79.93 & 83.41 & 84.94 & 87.32\\
        \hline
    \end{tabular}
    \label{tab:pruning_cifar}
\end{table*}

We compare the behaviour of the loss function (left) and its smoothed version (right) for each of the experts at training, where the experts are added sequentially as new tasks arrive. The results demonstrate that the smoothed loss more effectively reduces short-term variations and emphasizes long-term patterns. For more details, please see figure in the supplementary materials.

In Figures~\ref{fig:tasks-a}-\ref{fig:tasks-c} we illustrate the behaviour of average accuracy while model is trained on the sequence of tasks. The proposed algorithm, \tame, has the least drop in performance when adding more tasks among all considered methods and datasets.

In Figure~\ref{fig:cap} we demonstrate the effect of the buffer capacity on the accuracy of the selector network for Split CIFAR-100(20) data set. Note that the selector network accuracy depends on the similarity of tasks. For instance if we use $20$ super-classes from Split CIFAR-100(20), where similar labels are grouped together, the selector accuracy increases from $\sim62\%$ to $\sim79\%$.  

We also show the effect of the buffer capacity for pruning in Table \ref{tab:pruning_cifar}. For Permuted MNIST and Split MNIST, even a small size of buffer yield a good average accuracy.

\begin{figure*}
  \centering
  \begin{subfigure}{0.48\linewidth}
   \includegraphics[width=\linewidth] {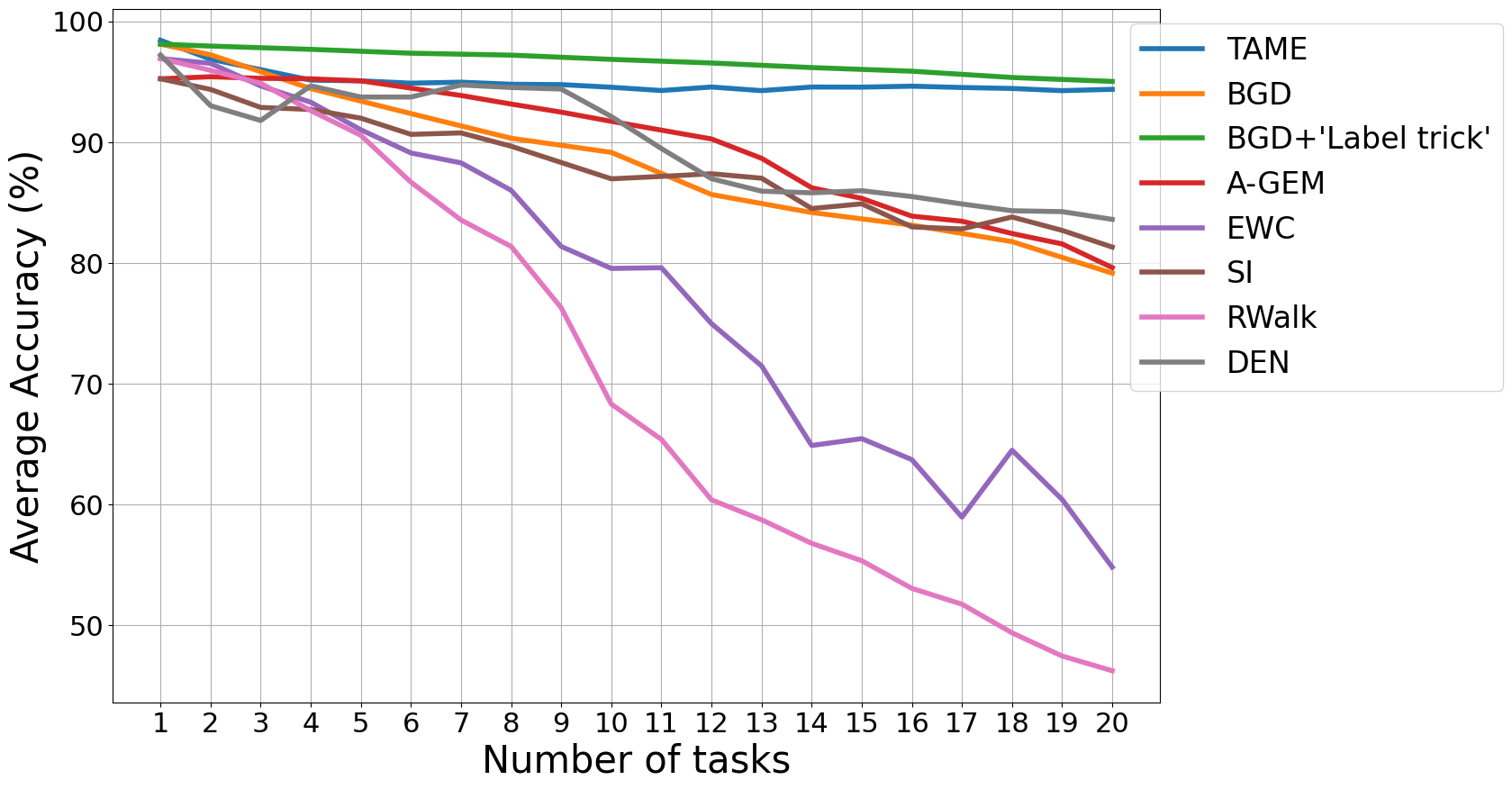}
    \caption{Permuted MNIST}
    \label{fig:tasks-a}
  \end{subfigure}
  \hfill
  \begin{subfigure}{0.48\linewidth}
    \includegraphics[width=\linewidth]{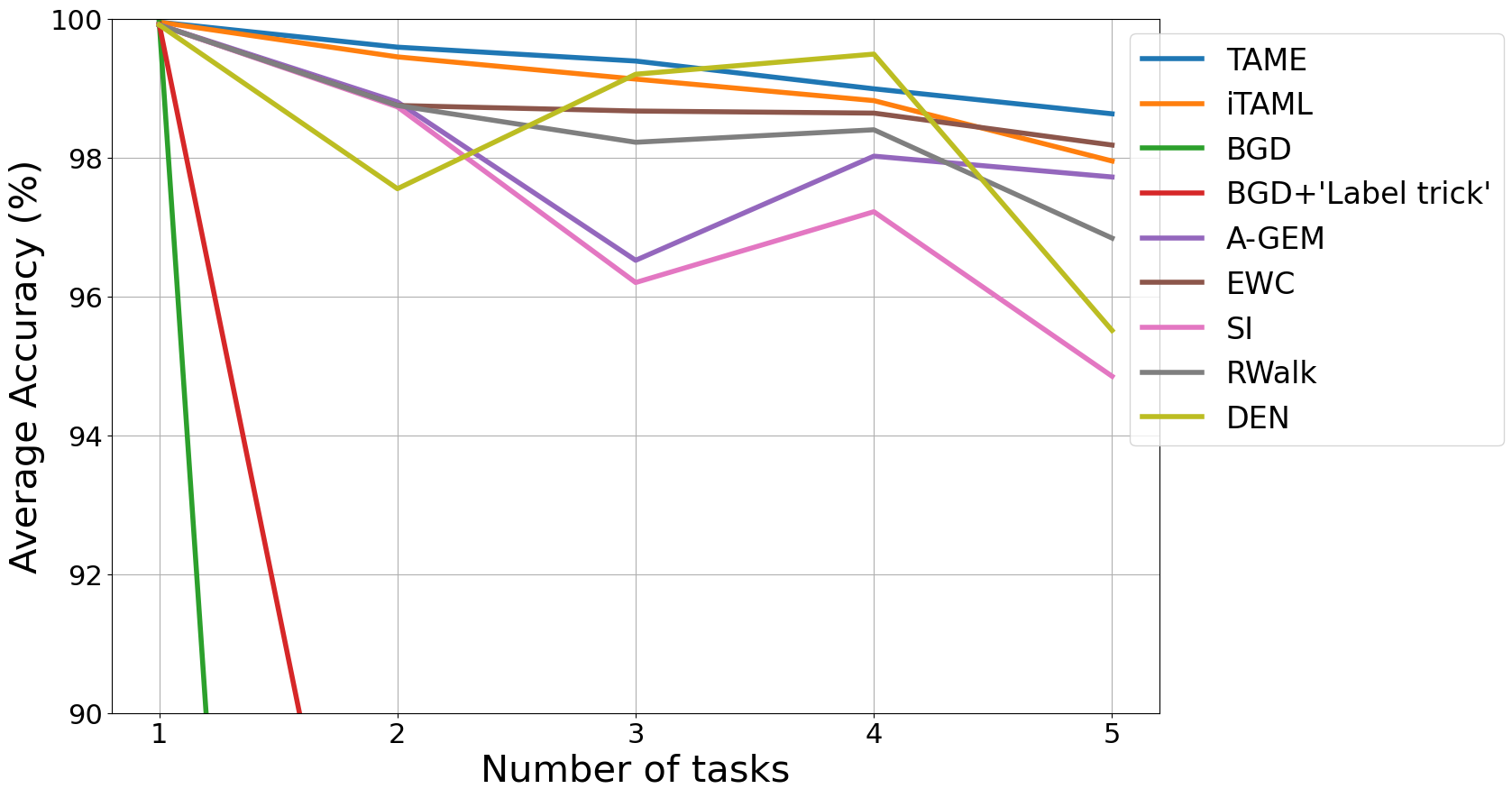}
    \caption{Split MNIST}
    \label{fig:tasks-b}
  \end{subfigure}
  \vfill
  \begin{subfigure}{0.48\linewidth}
   \includegraphics[width=\linewidth]{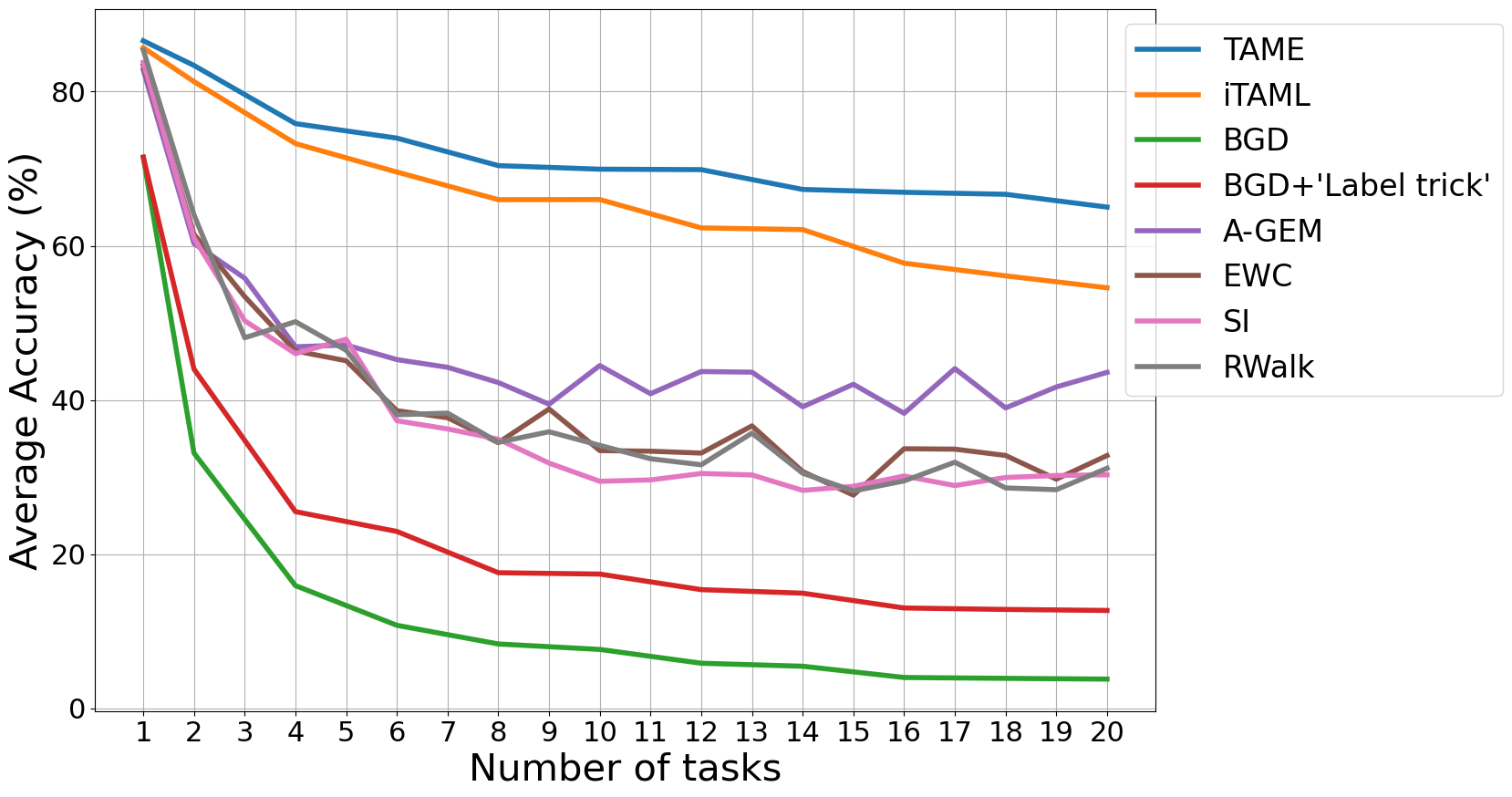}
    \caption{Split CIFAR-100}
    \label{fig:tasks-c}
  \end{subfigure}
  \hfill
  \begin{subfigure}{0.48\linewidth}
    \includegraphics[width=\linewidth]{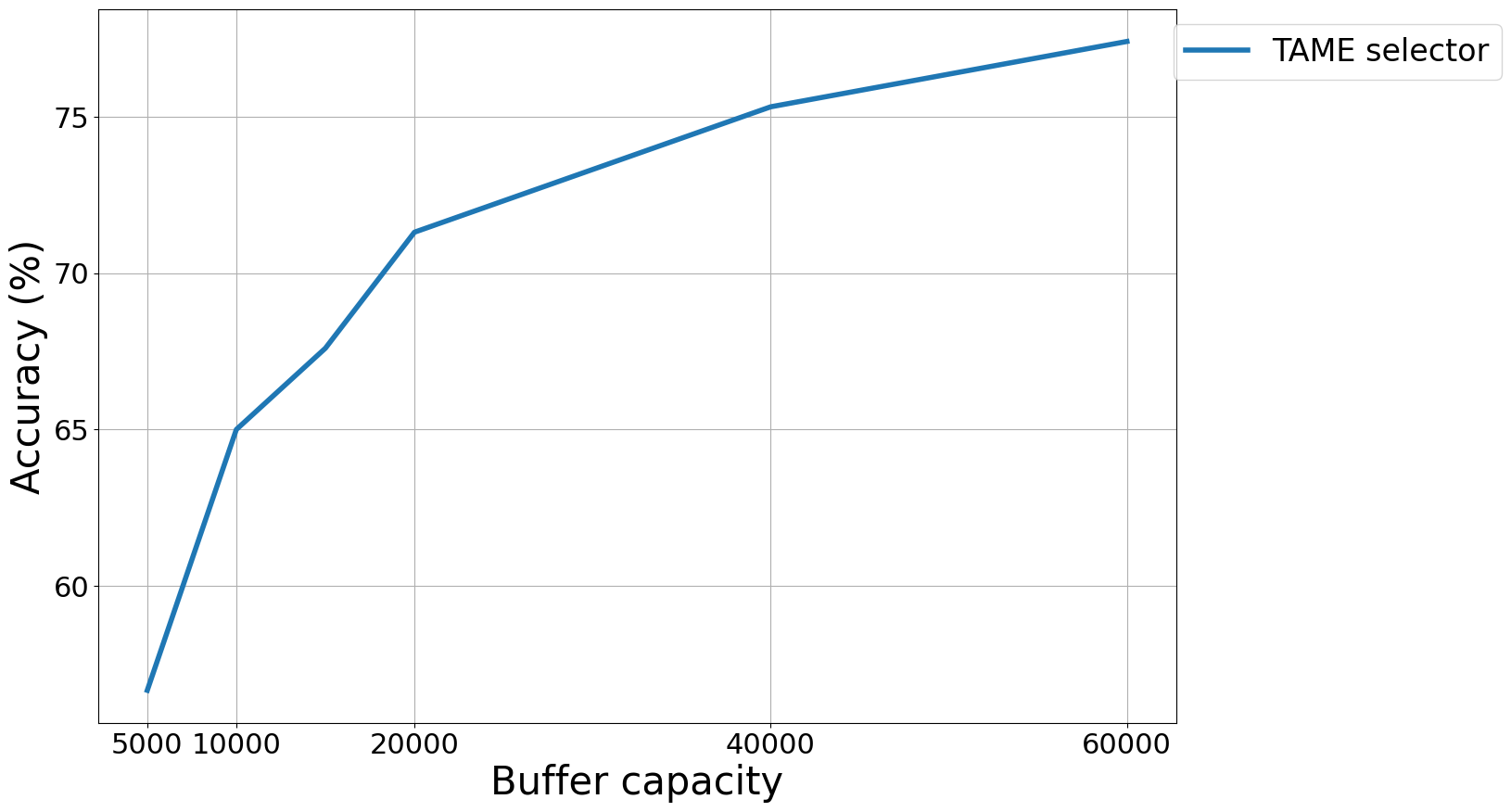}
    \caption{\tame selector, Split CIFAR-100}
    \label{fig:cap}
  \end{subfigure}
  \caption{({\bf a, b, c}) Average Accuracy versus the number of tasks for various data sets.
({\bf d}) Effect of the buffer capacity on the accuracy of the selector network in \tame for Split CIFAR-100 data set for 20 tasks.
}
  \label{fig:tasks}
\end{figure*}

\begin{figure*}
  \centering
  \begin{subfigure}{0.48\linewidth}
    \includegraphics[width=\linewidth]{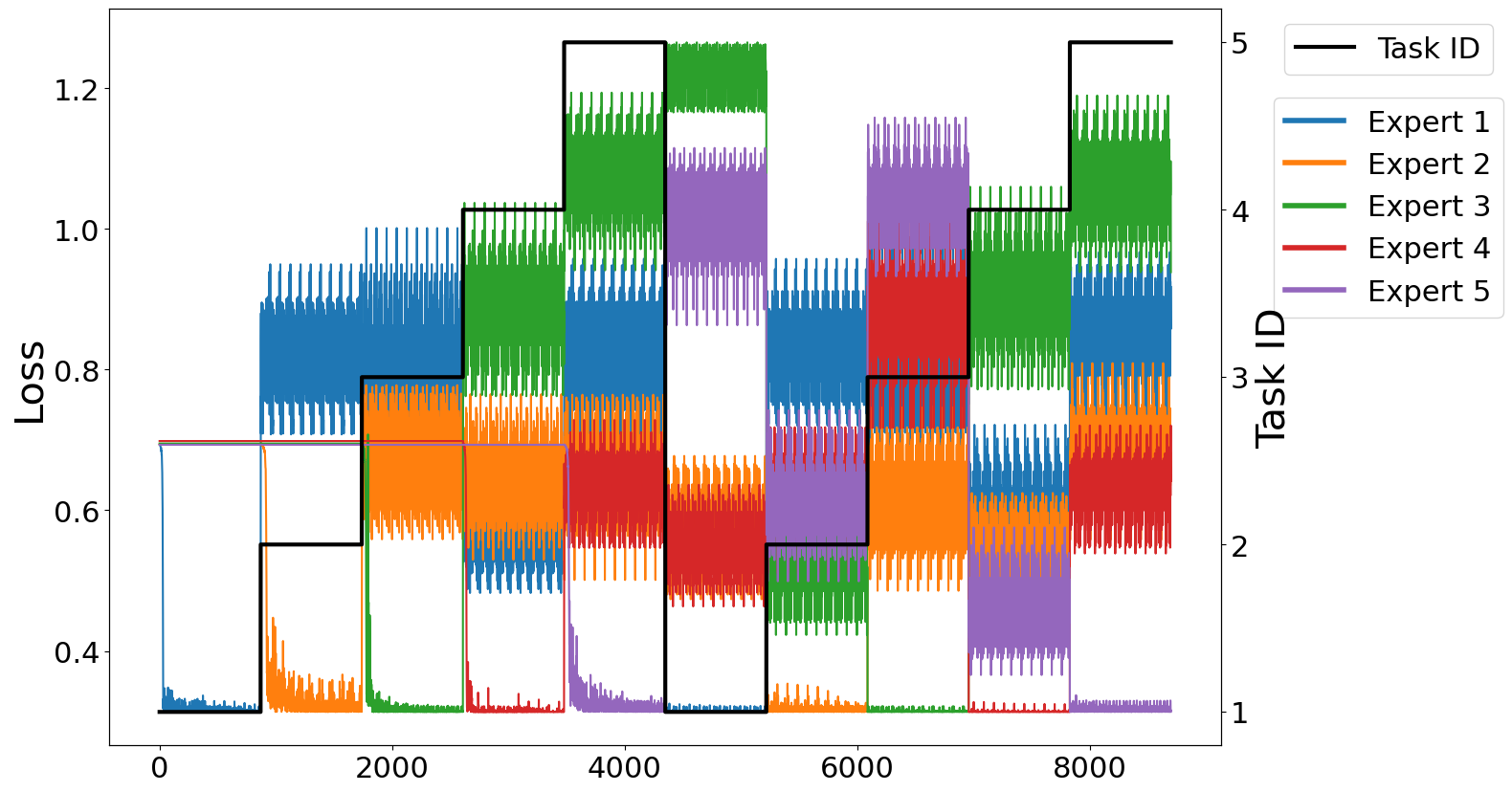}
    \caption{Loss value vs. iterations}
    \label{fig:switch-a}
  \end{subfigure}
  \hfill
  \begin{subfigure}{0.48\linewidth}
   \includegraphics[width=\linewidth]{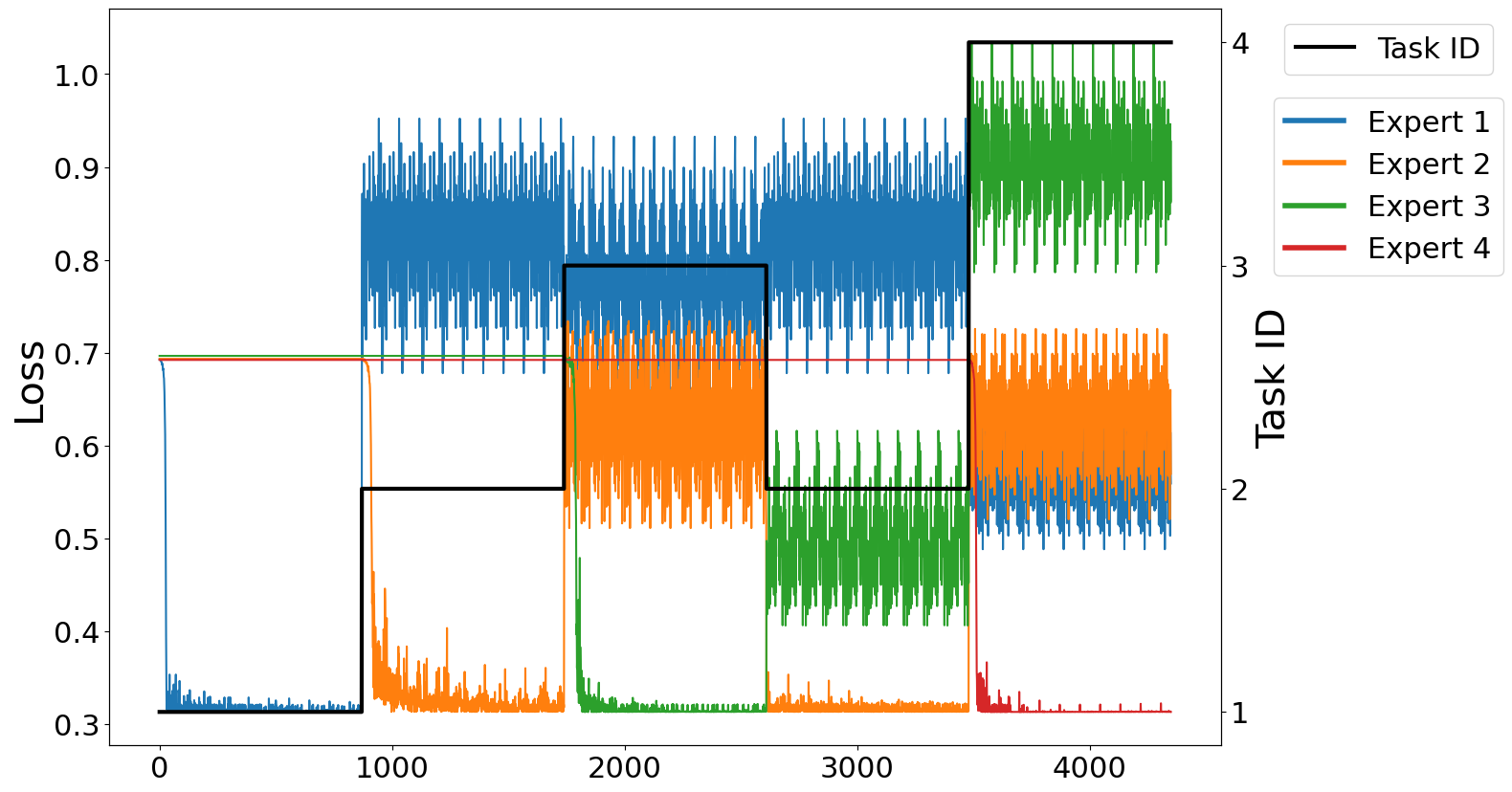}
    \caption{Loss value vs. iterations}
    \label{fig:switch-b}
  \end{subfigure}
  \caption{The value of loss function of different task expert networks during training on Split MNIST data set. Shown for two sequences of tasks: ({\bf a}) $\mathcal{T} = \{t_1, \dots, t_5, t_1,\dots,t_5\}$ ({\bf b}) $\mathcal{T} = \{t_1, t_2, t_3, t_2, t_4\}$. The algorithm switches to existing experts when a previously seen task occurs later in the sequence}
  \label{fig:expert-switch}
\end{figure*}

We also present in Figure~\ref{fig:expert-switch} the ability of the model to use the existing experts when a previously seen task appears again in the sequence. For instance in Figure~\ref{fig:switch-b} when task 2 appears for a second time in the sequence of $\mathcal{T} = \{t_1, t_2, t_3, t_2, t_4\}$ the algorithm appropriately switches to the already existing expert 2 rather than instantiating a new expert.

\section{Conclusion and Discussion}
\label{sec:con}
This paper addresses a more challenging scenario of continual learning - a task-agnostic setting, where the model is not provided with task descriptors during training or testing time. We devise a new continual learning algorithm for this purpose, that we call \tame, which is based on multiple expert networks associated with various tasks. These expert networks are added sequentially to the model in an online manner. During training, the algorithm automatically detects the task-switches based on statistically-significant deviation in the values of the loss function. At testing, the task identity is estimated by a selector network that is trained on a subset of training data that was drawn uniformly at random from all tasks. 
Experimental results show the efficacy of our approach on standard continual learning data sets, outperforming previous state-of-the-art techniques in terms of performance and model size. Specifically, we outperform the previous task-agnostic methods BGD, iTAML, HCL, and CN-DPM on various data sets, as well as the other techniques that take advantage of the knowledge of task descriptors at least during training.

{
    \small
    \bibliographystyle{ieeenat_fullname}
    \bibliography{main}
}

\clearpage
\setcounter{page}{1}
\maketitlesupplementary

\begin{figure*}[ht!]
  \centering
  \begin{subfigure}{0.48\linewidth}         {\label{fig:a_permuted}\includegraphics[width=\linewidth]{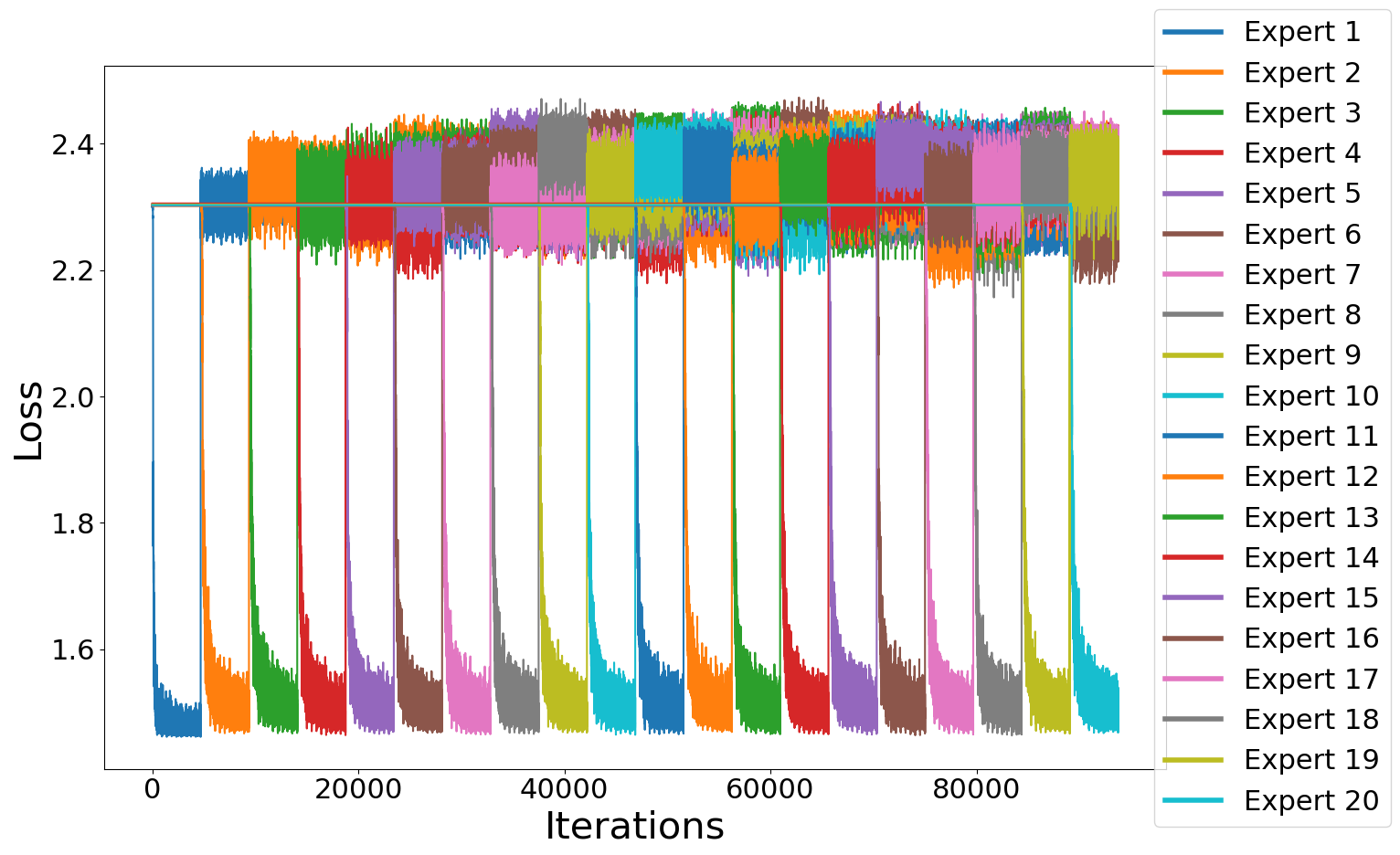}}
    \caption{Loss, Permuted MNIST}
  \end{subfigure}
  \hfill
  \begin{subfigure}{0.48\linewidth}
    {\label{fig:b_permuted}\includegraphics[width=\linewidth]{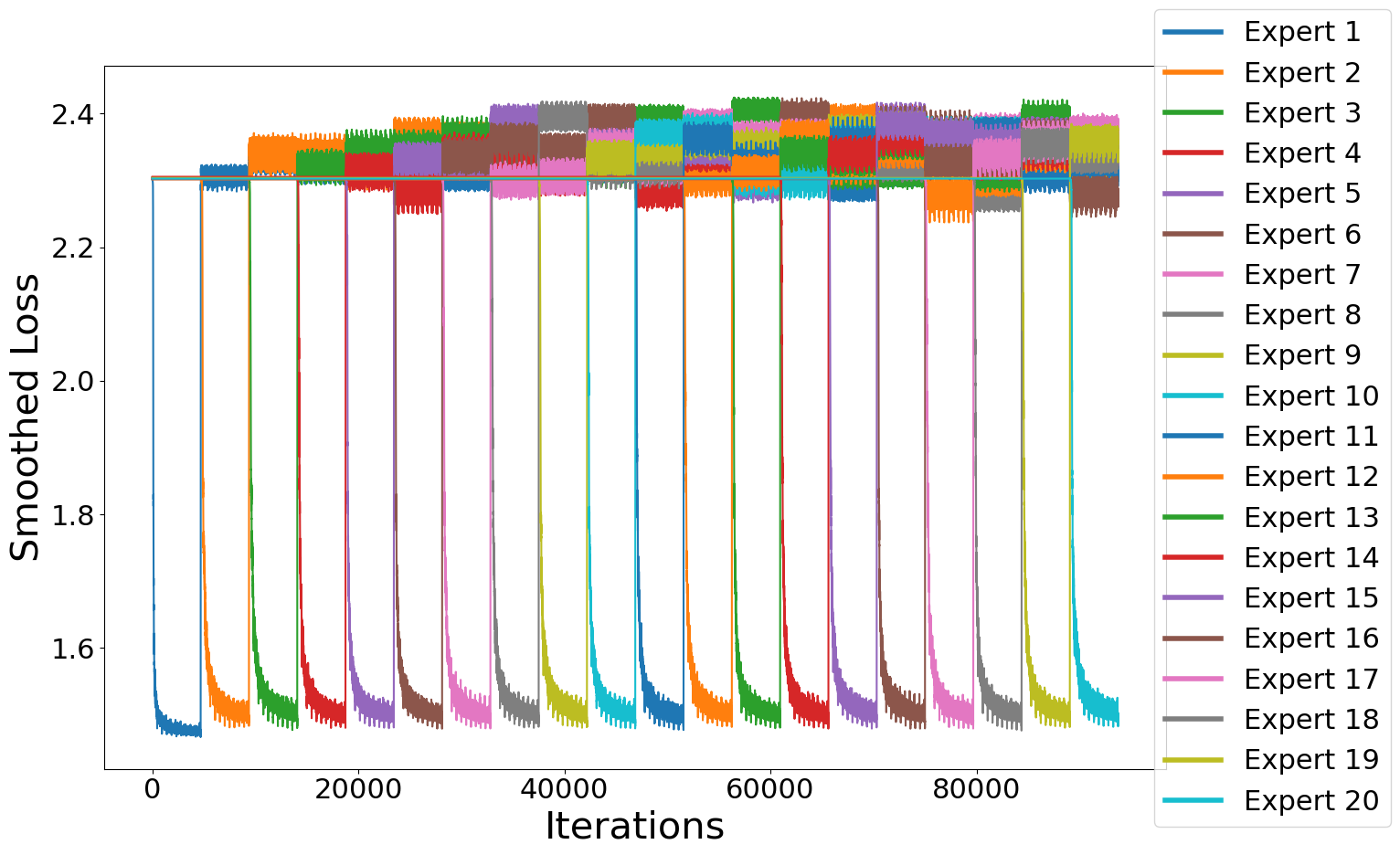}}
    \caption{Smoothed loss, Permuted MNIST}
  \end{subfigure}
  \vfill
  \begin{subfigure}{0.48\linewidth}
    {\label{fig:a_split}\includegraphics[width=\linewidth]{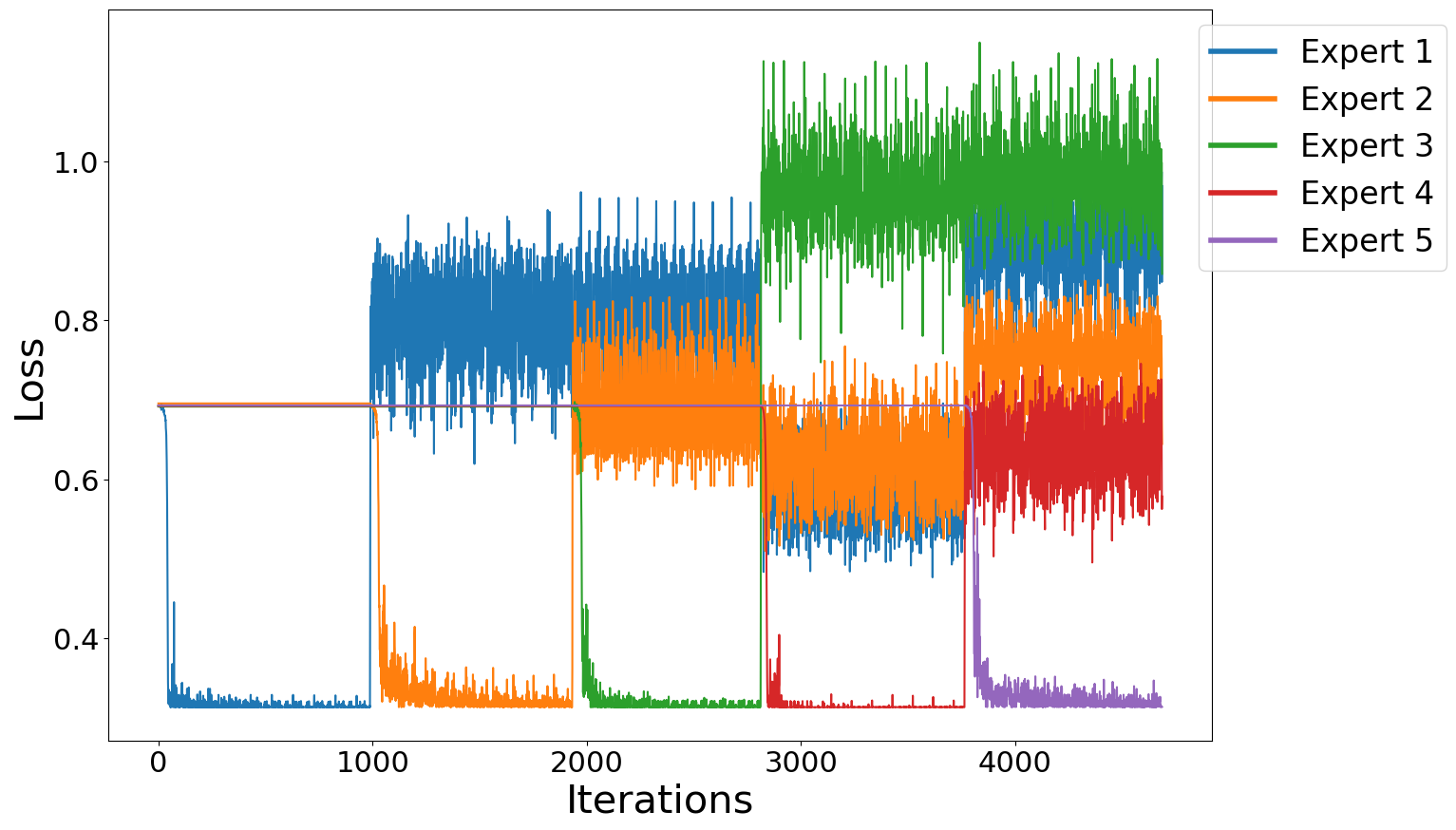}}
    \caption{Loss, Split MNIST}
    
  \end{subfigure}
  \hfill
  \begin{subfigure}{0.48\linewidth}
    {\label{fig:b_split}\includegraphics[width=\linewidth]{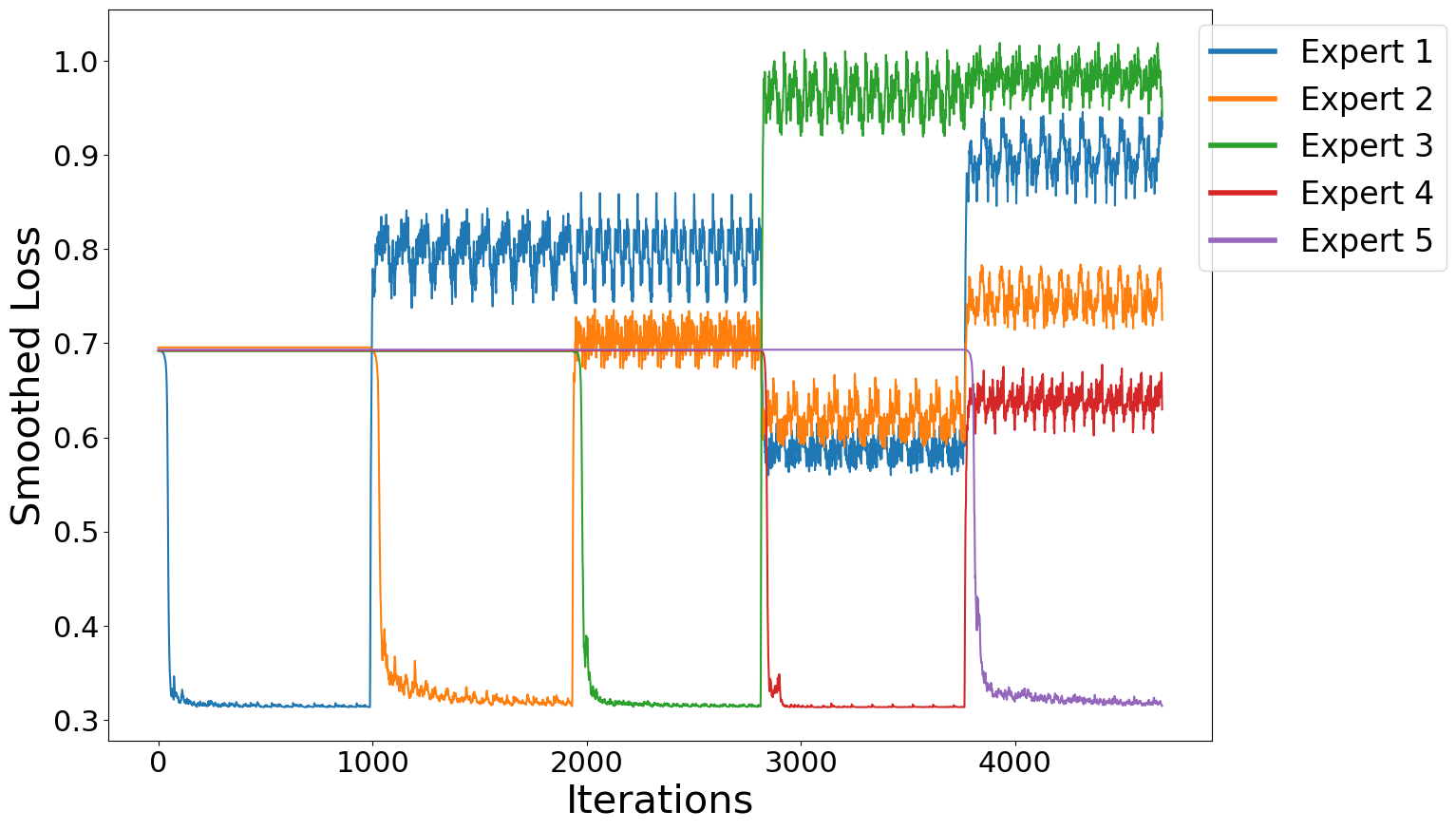}}
    \caption{Smoothed loss, Split MNIST}
    
  \end{subfigure}
  \vfill
  \begin{subfigure}{0.48\linewidth}
    {\label{fig:a_cifar}\includegraphics[width=\linewidth]{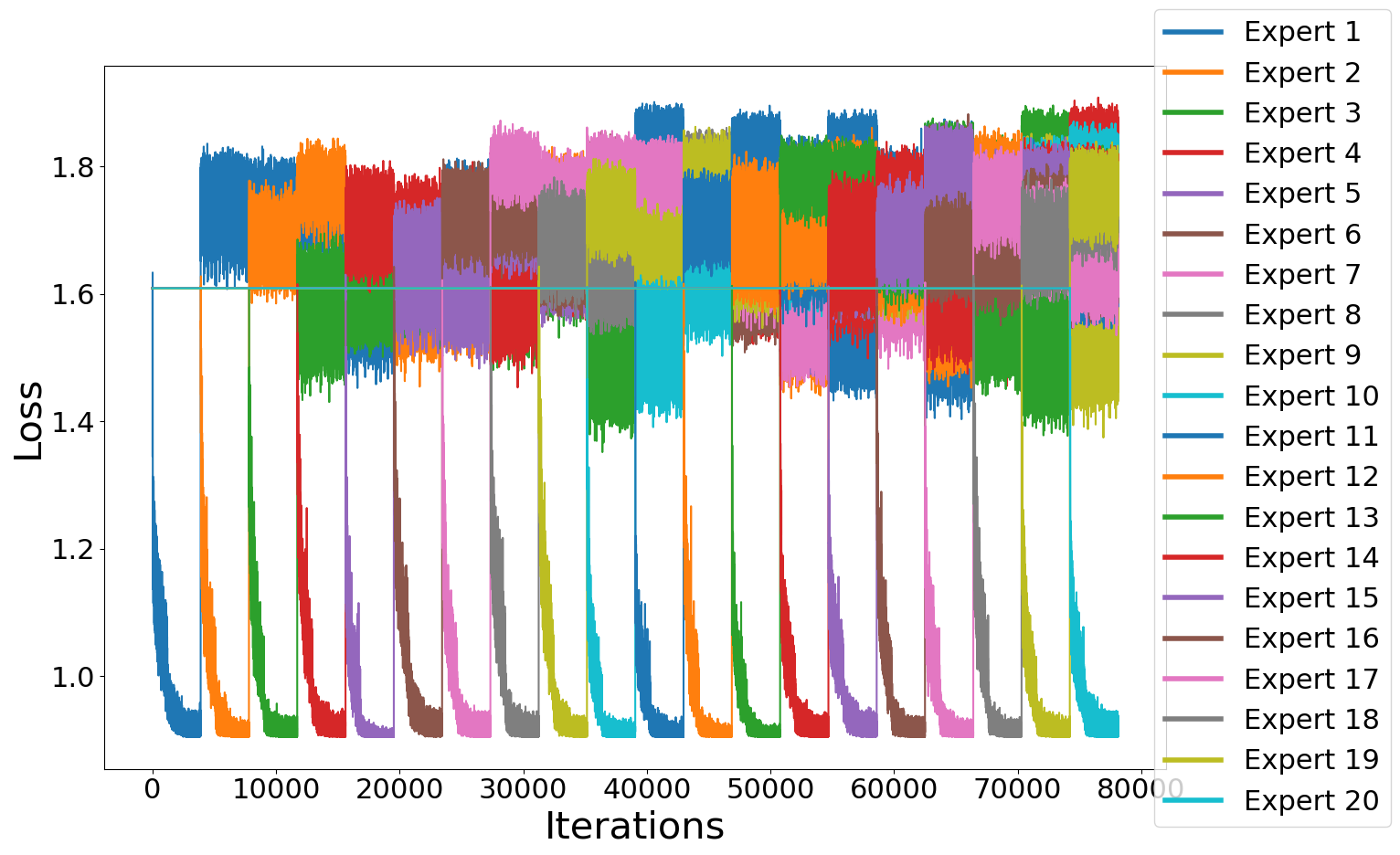}}
    \caption{Loss, Split CIFAR-100 (20)}
   
  \end{subfigure}
  \hfill
  \begin{subfigure}{0.48\linewidth}
   {\label{fig:b_cifar}\includegraphics[width=\linewidth]{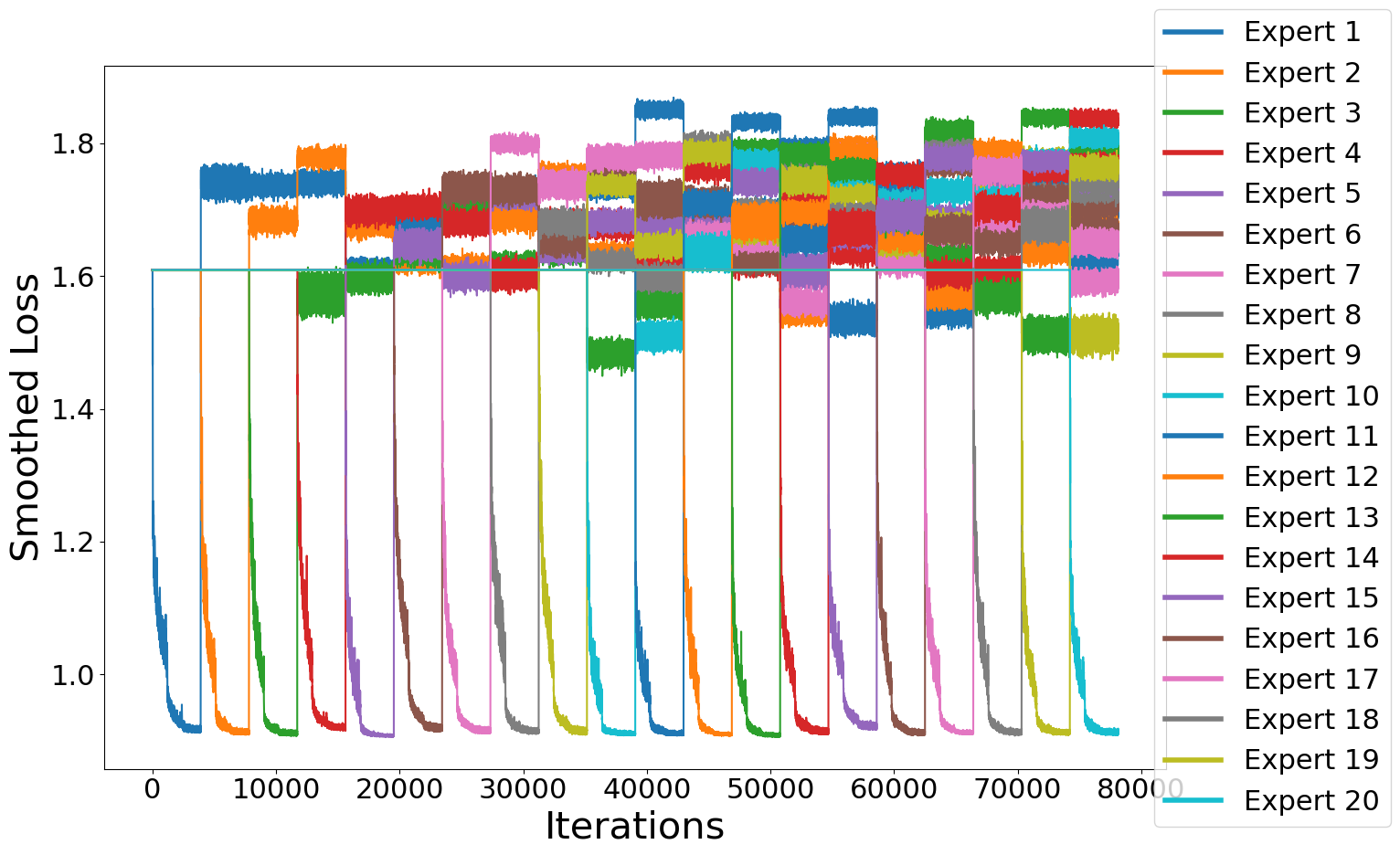}}
    \caption{Smoothed loss, Split CIFAR-100 (20)}
   
  \end{subfigure}
  \caption{
The value of loss function of different task expert networks (left) and its smoothed versions (right) during training for various data sets. Note that the experts are added sequentially as new tasks arrive. In all plots there is a clearly visible loss increase when a new task arrives
\label{fig:expert-loss}}
\end{figure*}

\begin{table}[ht!]
    \centering
    \caption{Hyperparameter search for the regularization-based techniques}
    \scalebox{1.0}{
    \begin{tabular}{|p{25mm}||p{30mm}|p{30mm}|p{35mm}|}
    \hline \text { Name } & \text { Permuted MNIST } & \text { Split MNIST } & \text { Split CIFAR-100 (20)} \\
    \hline \text { EWC } ($\lambda$: memory strength) & $\{10,20,50,100\}$ & $\left\{10^{4}, 10^{5}, 10^{6}, 10^{7}\right\}$ & $\left\{10^{4}, 10^{5}, 10^{6}, 10^{7}\right\}$\\
    \hline \text {SI} ($c$: dimensionless strength ) &  $\{0.01,0.1,1,10\}$ &  $\{0.01,0.1,1,100\}$ &  $\{0.01,0.1,1,10\}$ \\
    \hline \text { RWALK} ($\lambda$: regularization term) & $\{0.01,0.1,1,100\}$ &  $\{0.01,0.1,1,10\}$ & $\left\{10^{1}, 10^{2}, 10^{3}, 10^{4}\right\}$\\
    \hline
    \end{tabular}}
    \label{tab:hyper}
\end{table}

\begin{table}[ht!]
    \caption{Hyperparameter settings used for A-GEM}
    \centering
    \begin{tabular}{|c||c|c|c|}
    \hline \text { A-GEM } & \text { Permuted MNIST } & \text { Split MNIST } & \text { Split CIFAR-100  (20)} \\
    \hline \text {Episodic memory size } & 256 & 256 & 512 \\
    \hline \text { Episodic batch size } & 256 & 256 & 1300 \\
    \hline
    \end{tabular}
    \label{tab:hyper-agem}
\end{table}

\end{document}